# CORRELATION TRACKING VIA ROBUST REGION PROPOSALS

*Yuqi Han, Jinghong Nan, Zengshuo Zhang, Jingjing Wang and Baojun Zhao**

*Beijing Key Laboratory of Embedded Real-time Information Processing Technology*
*Beijing Institute of Technology, Beijing, China*
*Corresponding email:* zbj@bit.edu.cn

## Abstract

Recently, correlation filter-based trackers have received extensive attention due to their simplicity and superior speed. However, such trackers perform poorly when the target undergoes occlusion, viewpoint change or other challenging attributes due to pre-defined sampling strategy. To tackle these issues, in this paper, we propose an adaptive region proposal scheme to facilitate visual tracking. To be more specific, a novel tracking monitoring indicator is advocated to forecast tracking failure. Afterwards, we incorporate detection and scale proposals respectively, to recover from model drift as well as handle aspect ratio variation. We test the proposed algorithm on several challenging sequences, which have demonstrated that the proposed tracker performs favourably against state-of-the-art trackers.

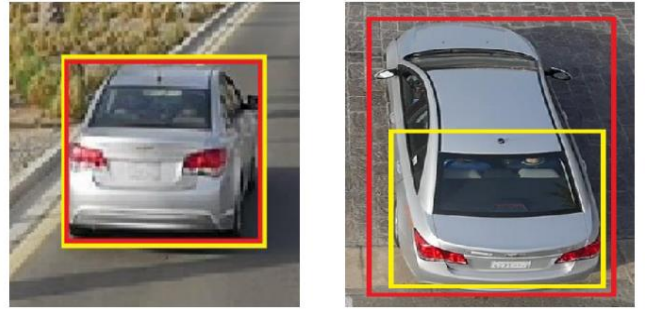

Fig. 1 The commonly existing ambiguity in visual tracking. With only the initial bounding box given (in the left), it's blurry to judge which tracking result (the red or yellow ones) is more accurate, as the viewpoint changes significantly.

## 1 Introduction

VISUAL tracking is one of the fundamental task in computer vision with a plethora of applications such as robotics, video surveillance, human-computer interaction etc. Recently, tracking approaches based on Correlation Filters (CF) [1–6] have received considerable attentions due to the high computational efficiency with the use of fast Fourier Transforms and their outstanding performance in public evaluation dataset and benchmark. Such approaches regress all the circular-shifted versions of input features to soft labels generated, producing less ambiguous response map than the traditional binary classifier, which show greater stability.

Despite of their huge success and development, such trackers still suffer from the model drift problem due to the limitation of tracking application itself. Unlike other tasks with clearly-defined target category, the goal of visual tracking is to estimate all kinds of targets' state and trajectory with only reliable information given at the initial frame. However, such information may be ambiguous or misleading under some circumstances. Fig.1 illustrates a vehicle tracking procedure. At the first frame, the car back is assigned to be target by the bounding box. As the sequence goes on, the target's pose varies significantly. It's hard to determine either the car back bounded in yellow or the whole car bounded in red is the better tracking result. Up to now, an explicit definition about the target to be tracked is still absent. After looking up several common online tracking datasets, we obtain an unwritten assumption that the tracking target is usually the whole object rather than its constitute parts.

Hence, the traditional CF trackers which adopt the template-matching approach is easy to overfit and drift gradually under challenging sequences due to the absence of target's prior knowledge. To tackle such limitation, in this paper, we propose an adaptive region proposal scheme to facilitate tracking. We firstly advocate a novel criterion to monitor tracking condition as well as determine potential failure. In addition, we show that generating a small number of high-quality candidate samples with the objectness information [7] taken into account, is impactful in recovering from tracking failure caused by the challenging attributes. For scale and aspect ratio estimation, the proposed tracker performs more effective than some existing scale-adaptive correlation-based tracking methods due to the flexible sampling manner and weaker assumption. Experiments on challenging sequences have demonstrated that the proposed algorithm performs favourably against the existing state-of-the-art trackers.

## 2. Proposed Method

In this section, we propose a two-stream tracking framework using the adaptive region proposal method. Firstly, we give a brief introduction of the baseline tracker [3]. Afterwards, we propose a tracking monitoring indicator to determine the tracking confidence each frame. When the tracking condition is desirable, the tracker would utilize scale proposals to handle scale and aspect ratio variations. On the other hand, if the confidence score is low, which indicates a tracking failure, we would generate detection proposals to search for the region that may contain losing target.

## 2.1 Kernelized Correlation Filter

The correlation filter tracker could be broken down into three components, namely training, detection and updating. In training section, KCF tracker would learn a filter **w**, which minimize the error between the training samples $\varphi(x_i)$ and regression labels $y_i$. The training goal could be formulated as:

$$\min_{\mathbf{w}} \sum_i (\varphi(x_i)w - y_i)^2 + \lambda ||\mathbf{w}|| \tag{1}$$

$\lambda$ is the regularization parameter that penalizes over-fitting. Eq. (1) could be solved directly in Fourier domain since a circulant matrix can be diagonalized using DFT matrix as:

$$\widehat{\mathbf{w}} = \frac{\hat{x} \odot \hat{y}}{\hat{x} \odot \hat{x}^* + \lambda} \tag{2}$$

Here, $\odot$ denotes Element-wise Product Operation, $\wedge$ and $*$ indicates Discrete Fourier Transform and Conjugate Operation.

Kernel trick is applied to obtain a more powerful filter in the case of the non-linear regression. Under such condition, w would be denoted by a linear combination of the training samples $\mathbf{w} = \sum_i \alpha_i \varphi(x_i)$, where $\alpha$ is the dual parameter of **w**. Then the optimization solution transfers under $\widehat{\boldsymbol{\alpha}}$ as:

$$\widehat{\boldsymbol{\alpha}} = \frac{\hat{y}}{\hat{k}^{xx} + \lambda} \tag{3}$$

After training process, the detection section is carried out on an image patch **z** in the newly coming frame within a $M \times N$ window, which is centred at the last target position. The response could be derived as follows:

$$f(\mathbf{z}) = \mathcal{F}^{-1}(\hat{k}^{xz} \odot \widehat{\alpha}) \tag{4}$$

$\mathcal{F}^{-1}$ denotes the Inverse DFT (IDFT) and $\hat{k}^{xz}$ is the so-called kernel correlation whose $i^{th}$ element is $k(z_i, x)$. Therefore, the position of the target at each frame could be determined by the maximum response $f(\mathbf{z})_{max}$. At last, in order to maintain the historical appearance of the target, linear interpolation is incorporated for updating the dual coefficients $\widehat{\alpha}$ and base sample template $\hat{x}$ with a fixed learning rate η as:

$$\widehat{\boldsymbol{x}}_t = (1-\eta)\widehat{\boldsymbol{x}}_{t-1} + \eta \widehat{\boldsymbol{x}}_t \tag{5}$$

$$\widehat{\boldsymbol{\alpha}}_t = (1-\eta)\widehat{\boldsymbol{\alpha}}_{t-1} + \eta \widehat{\boldsymbol{\alpha}}_t \tag{6}$$

## 2.2 Tracking Confidence Monitoring

Most of the traditional correlation filters-based tracker update their model in each frame or at a fixed interval. However, we argue that such strategy may introduce noise or background information when tracking is inaccurate. In this section, we advocate a novel criterion to evaluate the tracking results. To be more specific, we would consider the maximum value as well as the distribution of response map coinstantaneous.

Since the response map indicates features similarity between the target template and input samples across searching window. Under the ideal condition, there should only have one sharp peak in targets actual position. However, the response map may fluctuate due to surrounding distractors, temporal occlusion or other challenging factors. For example, in Fig.2 the response of the target is lower than the one of the background or distractors. In such cases, adopting the maximum searching strategy utilized in traditional KCF would lead to model drift. Hence, in this section we propose a novel criterion called average peak-sidelobe ratio (APSR) to evaluate the response map in order to reveal the tracking condition precisely as follow:

$$\text{APSR} = \frac{f(\mathbf{z})_{max} - f(\mathbf{z})_{min}}{\frac{1}{M \times N - 1}(\sum f(\mathbf{z}) - f(\mathbf{z})_{max})} \tag{7}$$

Here $f(\mathbf{z})_{max}$ and $f(\mathbf{z})_{mim}$ indicate the maximum and minimum value of the response map respectively. The denominator denotes the mean of the response map except the peak value, which is used to evaluate the side-lode. The APSR value would be large if there is only one sharp peak. We would record the peak value $f(\mathbf{z})_{max}$ and the APSR value for each frame and compare them with their historical average values as threshold to determine whether model drift occurs.

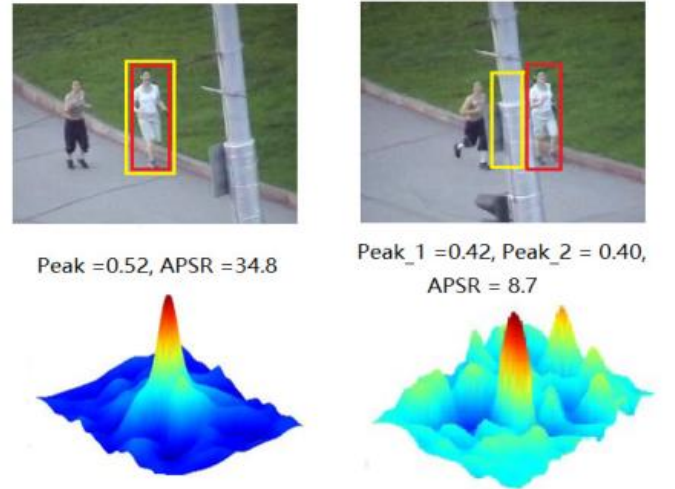

Fig. 2 shows the response map in Jogging sequence, where the response value of the distractor is competitive with the target. There KCF tracker would drift from the target due to its inflexible updating scheme. While with the use of the APSR criteria, whose value decreases rapidly when occlusion occurs. Our tracker could forecast the potential distraction and recover from model drift once the occlusion ends.

## 2.3 Integrating Objectness Proposals into Tracking

In this subsection, we employ re-detection and scale estimation based upon EdgeBox method [8] because of its fast speed and high recall. EdgeBox firstly computes an edge response for each pixel in the input image using a Structured Edge detector [9]. Then it traverses the whole image in a sliding window manner and scores every sampled box to select high-quality candidate samples. In [10], EdgeBox is applied as a post-processing step to improve tracker's adaptability to scale and aspect ratio change. In [11], Gao et al. employ EdgeBox method to facilitate the Struck tracker. Even though this is not the first time that introduce EdgeBox approach into tracking framework. It should be mentioned that these methods are substantially different from our work, where we integrate region proposal method into correlation tracking framework recurring to a novel monitoring criterion. Instead of only handling scale variation scene, the proposed algorithm could also address tracking failure issue as well.

As mentioned above, the tracker would calculate the value of $f(\mathbf{z})_{max}$ and APSR to determine whether tracking failure occurs. Once it happens, we would activate re-detection module by performing detection over several instance region-proposals, which cover the target. Instead of directly applying the computed high-scored proposals for tracking, we argue that the object instance level should be taken into account, since EdgeBox is a generic proposal generator, which may prone to generate false positive samples in a cluttered background. To this end, we incorporated an online updated SVM classifier to learn the target appearance as in [11]. But we update the SVM classifier using some generated instance-aware proposals as training data only if the tracking condition is ideal, which is decided by the monitoring criterion. We select the optimal detection results based upon the following objective function:

$$c^* = arg \max_{c_i \in B_t} f(c_i) + \zeta \times dist(P_{t-1}, P_{c_i}) \quad (8)$$

Here, $B_t = \{c_1, c_2, ...., c_n\}$ denotes candidate samples union generated by Edgebox method at frame t, $P_{t-1}$ indicates the center of the tracking box at last frame, $P_{c_i}$ is the center of the $i^{th}$ candidate sample at current frame, function $f(\cdot)$ is the response output between the template and candidate samples as introduced before. The second item denotes that taking motion constrain into consideration so as to reject model drift caused by distractors with similar appearance. Here, $\zeta$ is the penalty parameter to balance these two factors. We employ the same function to represent the motion constrain in [11] as:

$$dist(P_{t-1}, P_{c_i}) = \frac{1}{2b} exp(-\frac{1}{b}||P_{c_i} - P_{t-1}||) \quad (9)$$

Here, b is the diagonal length of the searching window size. It should be mentioned that the selected optimal position may vary abruptly and significantly. In order to maintain the consistency and robustness, we update the location $(x_t^{c^*}, y_t^{c^*})$ with a damping factor $\gamma_1$:

$$x_t = (1 - \gamma_1)x_{t-1} + \gamma_1 x_t^{c^*} \quad (10)$$

$$y_t = (1 - \gamma_1)y_{t-1} + \gamma_1 y_t^{c^*} \quad (11)$$

On the other hand, if the monitoring criterion indicates an ideal tracking circumstance, we would present how to tackle the scale variation and aspect ratio changes by incorporating scale proposals. Firstly, the tracker would generate numerous proposals centred at $P_{t-1}$. Secondly, we would sort them by their objectness score and pick up the top 200 candidates for further processing. Proposal rejection technique would be applied to filter the proposal whose intersection over union (IoU) with the bounding box is smaller than 0.6 or larger than 0.9. We consider the samples whose overlap rate exceed 0.9 are much like the current tracking results, while for the candidates whose IoU with the tracking results is lower than 0.6 are much likely to be false proposals. Afterwards, for proposals after rejection, we would resize those candidate samples with different size and aspect ratio into a fixed size (normally the template's size) and compute the response as Eq. (4) in spatial domain. Similar to detection procedure, we choose the optimal scale proposal as target's current size which yield the maximum response. Meanwhile, the updating strategy is same as the one used in re-detection procedure in order to guarantee the target size changes smoothly.

$$w_t = (1 - \gamma_2)w_{t-1} + \gamma_2 w_t^{c^*} \quad (12)$$

$$h_t = (1 - \gamma_2)h_{t-1} + \gamma_2 h_t^{c^*} \quad (13)$$

## 3 Experiments

To evaluate the effectiveness and robustness of our algorithm, we empirically validate the proposed tracker on 5 challenging sequences from Online Tracking Benchmark [12] with other state-of-the-art methods. These trackers could be broadly categorized into three classes: (1) baseline CF trackers including KCF [3] and DSST [5], (2) trackers using region proposal method or redetection module such as EBT [11] and LCT [6], (3) other representative trackers reported in Tracking Benchmark such as Struck [13] and TLD [14] methods.

### 3.1 Quantitative results

We evaluate all the trackers by adopting one common criteria: the overlap ratio. We denote the ratio $R = \frac{S(BT \cap BG)}{S(BT \cup BG)}$, where BT denotes the tracking results and BG is the ground-truth bounding box, R indicates the IoU of such boxes. The overlap ratio shows the percentage of frames with $R > t$, throughout all threshold $t \in [0,1]$. The average overlap rate is shown in Table 1. It demonstrates that our tracker outperforms other state-of-the-art methods in these sequences.

### 3.2 Qualitative results

Fig.3 illustrates qualitative results in sequences with challenging sequences compared with the other state-of-the-art trackers. Occlusion is a big challenge for visual tracking, as it would destroy the holistic appearance of the target. We test 2 sequences (David3, Jogging2) having severe occlusion. One can see that only LCT, EBT and our tracker could lock the target precisely due to the monitor indicator and detection module. It should be mentioned that even TLD is able to re-detect the target, it is sensitive to similar disturbance. In Dog1, the target undergoes large scale change. STRUCK, TLD and KCF couldn't adapt to such appearance variation. While DSST, LCT, EBT and our tracker could tail this challenging state for the entire video, which could be attributed to the scale searching strategy (scale pyramid or scale-instance proposals). However, for the sequences with viewpoint change and aspect ratio variation (Freeman3, CarScale), DSST and LCT tracker gradually drift from the target due to the inflexible pre-defined candidate sampling scheme. Moreover, even though EBT tracker could handle the aspect ratio change as our tracker, it could hardly recover from tracking failure caused by occlusion and fast appearance change. While with the objectness information, motion trajectory and tracking confidence all taken into consideration, our tracker could deal with the above issues, and performs better than other trackers.

Table 1 Average overlap rate. The red bold fonts and blue italic fonts indicate the best and the second-best performance.

|  | KCF | DSST | LCT | EBT | STRUCK | TLD | PROPOSED |
|---|---|---|---|---|---|---|---|
| David3 | 0.759 | 0.452 | 0.765 | *0.766* | 0.289 | 0.095 | **0.842** |
| Jogging2 | 0.122 | 0.139 | 0.714 | *0.765* | 0.199 | 0.647 | **0.941** |
| Dog1 | 0.548 | 0.544 | *0.813* | 0.679 | 0.543 | 0.593 | **0.868** |
| Freeman3 | 0.328 | 0.344 | 0.318 | 0.314 | 0.264 | *0.442* | **0.786** |
| CarScale | 0.412 | 0.439 | *0.674* | 0.578 | 0.415 | 0.452 | **0.749** |

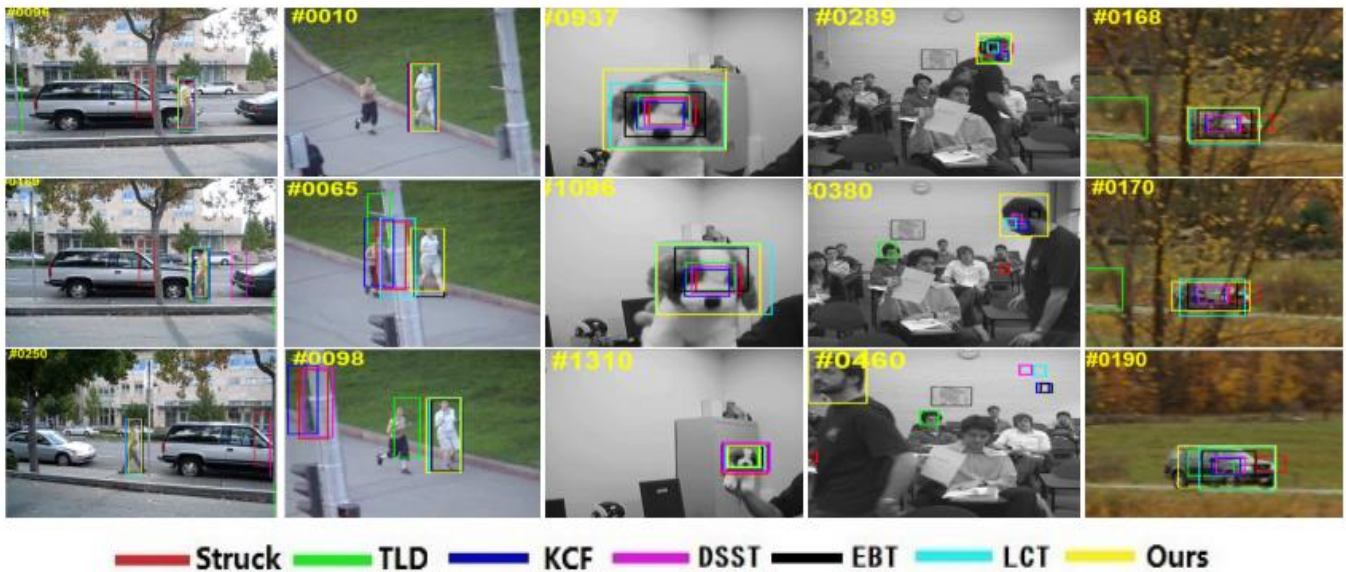

Fig. 3 Representative tracking results on challenging sequences.

## 4 Conclusion

In this paper, an effective two-stream framework is presented to enable tracking condition monitoring, failure redetection as well as scale adaptation. Specifically, we employ the region proposal technique, which could generate few yet high-quality candidate samples, into the well-known correlation tracking approach. With the consideration of tracking confidence and target objectness information, the proposed tracker performs favourably against other state-of-the-art trackers.

## 5 Acknowledgements

This work is supported by the Chang Jiang Scholars Programme (Grant No. T2012122), together with 111 Project of China under Grant B14010.